\documentclass{article}

\usepackage[preprint]{neurips_2025}
\usepackage{amsmath,amsfonts}
\usepackage{algorithmic}
\usepackage{array}
\usepackage[caption=false,font=normalsize,labelfont=sf,textfont=sf]{subfig}
\usepackage{textcomp}
\usepackage{stfloats}
\usepackage{url}
\usepackage{verbatim}
\usepackage{graphicx}
\usepackage{tabularx}
\usepackage{pifont}
\usepackage{rotating}
\usepackage{hyperref}
\usepackage{xurl}  
\usepackage{hyperref}
\usepackage {wrapfig}
\usepackage{tikz}
\usepackage{multirow} 
\usepackage{booktabs}
\usepackage{adjustbox}
\usepackage{xcolor}

\definecolor{lightgreen}{rgb}{0.56, 0.93, 0.56}
\definecolor{brightlavender}{rgb}{0.75, 0.58, 0.89}
\definecolor{capri}{rgb}{0.0, 0.75, 1.0}
\definecolor{darkpastelgreen}{rgb}{0.01, 0.75, 0.24}

\definecolor{tree-level-1}{RGB}{245,20,85}
\definecolor{tree-level-2}{RGB}{246,86,118}
\definecolor{tree-level-3}{RGB}{248,177,193}
\definecolor{tree-leaf}{RGB}{176,230,198}
\usepackage{dsfont}

\definecolor{titlecolor}{RGB}{0,0,0}
\definecolor{hidden-draw}{RGB}{0,0,0}
\definecolor{lighttealblue}{RGB}{0,0,0}    
\definecolor{lightplum}{RGB}{0, 0, 0}        
\definecolor{harvestgold}{RGB}{0, 0, 0}   

\usepackage{booktabs} 
\usepackage{colortbl} 

\definecolor{my_green}{RGB}{51,102,0}
\definecolor{my_red}{RGB}{204, 0, 0}
\newcommand{\eg}{\textit{e}.\textit{g}.}
\newcommand{\etal}{\textit{et al}.}

\newcommand{\etc}{\textit{etc}}

\usepackage{natbib}
\setcitestyle{numbers,square}

\usepackage[utf8]{inputenc} 
\usepackage[T1]{fontenc}    
\usepackage{hyperref}       
\usepackage{url}            
\usepackage{booktabs}       
\usepackage{amsfonts}       
\usepackage{nicefrac}       
\usepackage{microtype}      
\usepackage{xcolor}         
\usepackage{pifont}

\title{MLLMs are Deeply Affected by Modality Bias
}
\author{Xu Zheng$^{1,3,}$\thanks{These authors have equal contributions.}\quad Chenfei Liao$^{1,}$\footnotemark[1]\quad Yuqian Fu$^{3}$\quad Kaiyu Lei$^{4}$\quad Yuanhuiyi Lyu$^{1}$\quad \\ \textbf{Lutao Jiang$^{1}$\quad Bin Ren$^{5,6,3}$\quad Jialei Chen$^{7}$\quad Jiawen Wang$^{8}$\quad Chengxin Li$^{9,10}$}\quad \\ \textbf{Linfeng Zhang$^{11}$\quad Danda Pani Paudel$^{3}$\quad Xuanjing Huang$^{12}$\quad Yu-Gang Jiang$^{12}$}\quad \\ \textbf{Nicu Sebe$^{6}$\quad Dacheng Tao$^{13}$\quad Luc Van Gool$^{3}$\quad Xuming Hu$^{1,2,}$\thanks{Corresponding author.}} 
\\
$^{1}$HKUST(GZ)\quad $^{2}$CSE, HKUST \quad $^{3}$INSAIT, Sofia University “St. Kliment Ohridski”\quad \\ $^{4}$Xi'an Jiaotong University\quad $^{5}$University of Pisa, IT\quad $^{6}$University of Trento, IT\quad \\ $^{7}$Nagoya University\quad $^{8}$China University of Mining \& Technology, Beijing\quad \\ $^{9}$Tongji University\quad $^{10}$SPIC Energy Science and Technology Research Institute\quad \\ $^{11}$Shanghai Jiao Tong University \quad $^{12}$Fudan University\quad \\$^{13}$College of Computing \& Data Science, Nanyang Technological University
}

\begin{document}

\maketitle

\begin{abstract}
Recent advances in Multimodal Large Language Models (MLLMs) have shown promising results in integrating diverse modalities such as texts and images. 
MLLMs are heavily influenced by modality bias, often relying on language while under-utilizing other modalities like visual inputs.
This position paper \textbf{\textit{argues that MLLMs are deeply affected by modality bias}}. Firstly, we diagnose the current state of modality bias, highlighting its manifestations across various tasks. Secondly, we propose a systematic research road-map related to modality bias in MLLMs. Thirdly, we identify key factors of modality bias in MLLMs and offer actionable suggestions for future research to mitigate it. To substantiate these findings, we conduct experiments that demonstrate the influence of each factor:
\ding{172} Data Characteristics: Language data is compact and abstract, while visual data is redundant and complex, creating an inherent imbalance in learning dynamics. 
\ding{173} Imbalanced Backbone Capabilities: The dominance of pretrained language models in MLLMs leads to overreliance on language and neglect of visual information.
\ding{174} Training Objectives: Current objectives often fail to promote balanced cross-modal alignment, resulting in shortcut learning biased toward language.
These findings highlight the need for balanced training strategies and model architectures to better integrate multiple modalities in MLLMs. We call for interdisciplinary efforts to tackle these challenges and drive innovation in MLLM research.
Our work provides a fresh perspective on modality bias in MLLMs and offers insights for developing more robust and generalizable multimodal systems—advancing progress toward Artificial General Intelligence.
\end{abstract}

\section{Introduction}
\subsection{Background}
Multimodal Large Language Models (MLLMs)~\cite{bai2025qwen2,wang2024qwen2,zhu2025internvl3,chen2024expanding,team2024gemini,hurst2024gpt} have revolutionized the ability to handle diverse modalities, including text, image, audio, video, and other emerging modalities (tactile~\cite{dahiya2009tactile,zou2017novel,chi2018recent}, event~\cite{gallego2020event,zheng2023deep,rebecq2019high}, panoramic image~\cite{zheng2024360sfuda++,zheng2024semantics,zhong2025omnisam}, \etc). This expansion into the multimodal domain typically involves pretraining with multimodal data pairs or fine-tuning on specialized multimodal instruction datasets~\cite{liu2024mmbench,fang2024mmbench,mathew2021docvqa,li2024survey,li2024survey1}. MLLMs excel at understanding complex multimodal patterns and translating them into a coherent language representation space~\cite{wu2024visual,jiang2025survey,huo2025continue}. Despite significant advancements, challenges remain, where one major issue is \textbf{"Modality Bias"}. As in ~\cite{zhang2024debiasing}, MLLMs often generate content that is disproportionately influenced by the underlying language model used during pretraining, rather than the input images or other modalities. In cases where images are noisy or even absent, MLLMs still confidently generate answers, highlighting a clear bias towards learned language patterns over multimodal integration~\cite{park2025assessing,tong2024eyes}.

An ideal MLLM should be modality-balanced, effectively integrating useful information from all modalities to provide reliable, accurate, and comprehensive answers~\cite{chen2024we}. Achieving this balance is crucial for overcoming modality bias and ensuring that the model can leverage the full potential of each modality in multimodal tasks~\cite{chen2024quantifying,yue2024mmmu}. 

\begin{figure}
    \centering
    \includegraphics[width=1\linewidth]{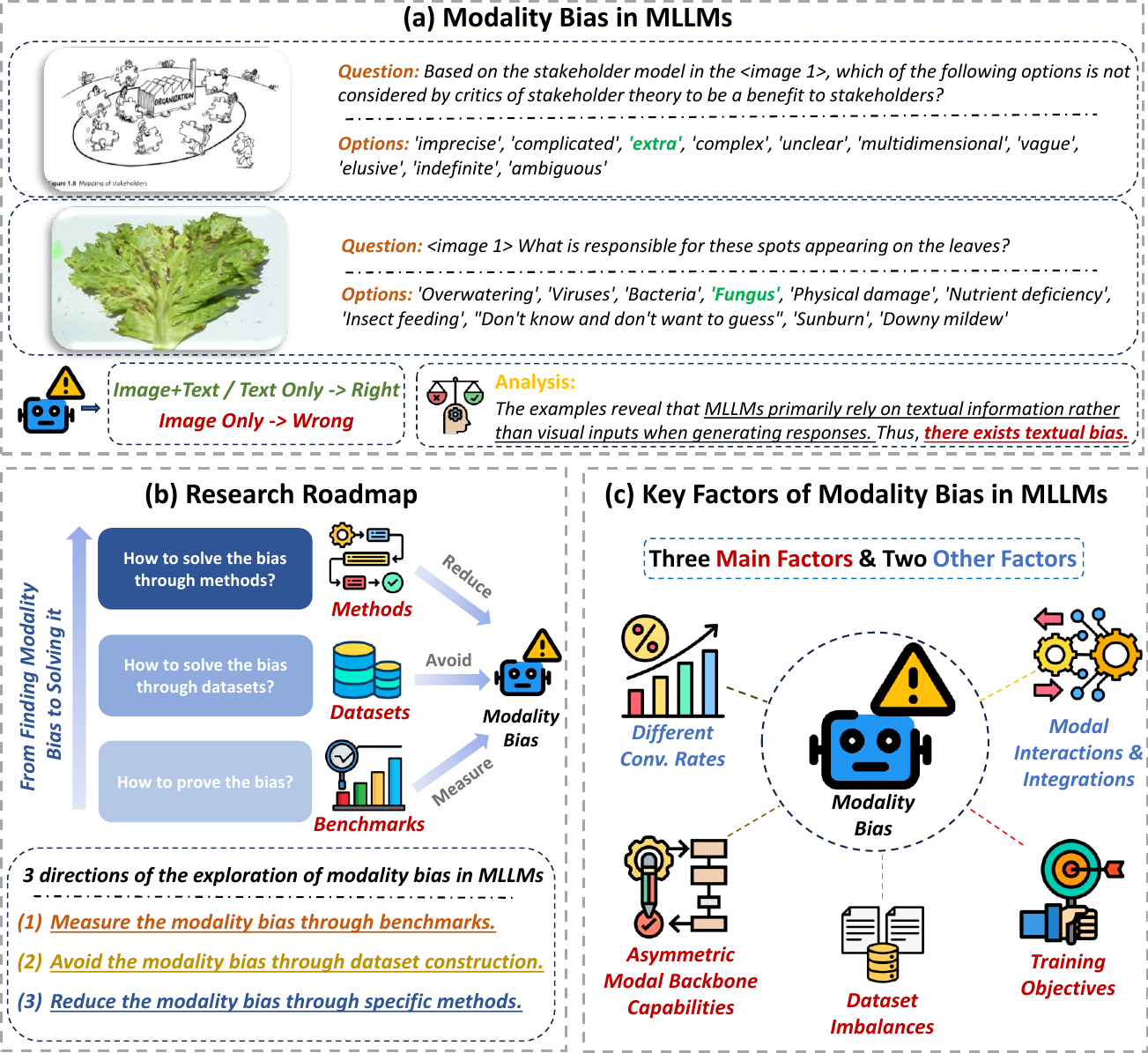}
    \caption{The big picture of our position on modality bias in MLLMs.
(a) We define modality bias in multimodal models, with a focus on MLLMs.
(b) We outline a research roadmap on modality bias, highlighting three key directions.
(c) We summarize five contributing factors to modality bias: three primary and two secondary.}
    \label{fig:big}
\end{figure}

\subsection{Modality Bias Phenomenon}

Multimodal learning improves neural networks' cross-modal comprehension by fusing heterogeneous data modalities (\eg, images, text, audio, and video), thereby enabling better world modeling~\cite{xu2023multimodal,bayoudh2022survey,jabeen2023review,zheng2024learning,liao2025memorysam,brodermann2025cafuser}. It is widely assumed that leveraging multiple input modalities will lead to improved model performance~\cite{manzoor2023multimodality}. However, research has demonstrated that these modalities are not always utilized to their full potential~\cite{wei2024enhancing,zheng2024centering,zheng2025reducing,zheng2024learning1}. Despite achieving superior performance over unimodal models, multimodal models still fail to fully exploit the capabilities of each modality~\cite{peng2022balanced}.

As noted by \cite{alabdulmohsinclip}, multimodal models, particularly those employing contrastive learning techniques such as CLIP~\cite{radford2021learning,tschannen2023clippo,yang2024clip,wu2023tinyclip,li2023inverse}, learn representations from multimodal data but may inadvertently inherit biases due to the imbalanced and biased nature of the training data. As discussed in \cite{xu2025balancebenchmark}, this issue severely hinders the effectiveness of multimodal learning. In detail, such a condition occurs when certain modalities dominate the training process while others remain underrepresented, constraining the model’s capacity to capture the comprehensive information embedded in multimodal data distributions. Such dominance can lead to models' over-reliance on the dominant modality, which impedes their ability to generalize effectively to unseen data or situations where the dominant modality is absent.

Moreover, in multimodal systems, modality bias can manifest in several other ways. For instance, if a model is primarily trained on image-text pairs, but audio or video data is only sparsely represented, the model may learn representations that are disproportionately influenced by the image-text modalities, while neglecting the rich information available from the audio or video inputs. This type of imbalance leads to a model that may perform well under normal circumstances but struggles to generalize when the dominant modality is absent.

From a model learning perspective, \cite{yang2024facilitating} identifies the differing convergence rates of modalities as a core cause of modality bias. The varying levels of difficulty in fitting category labels across different modalities contribute to this disparity. Some modalities may align more easily with the target labels than others, leading to an unequal contribution to the final learned representations. This uneven convergence further exacerbates the problem, reinforcing the bias towards certain modalities.

Recent studies in multimodal scene understanding have further highlighted this issue. For example, research by \cite{zheng2024centering}, \cite{zheng2024learning}, and \cite{liao2025benchmarking} shows that multimodal segmenters often over-rely on certain modalities, resulting in significant performance degradation when these dominant modalities are missing or unavailable. These findings underscore the need to address modality imbalance in order to ensure that all modalities contribute effectively to the learning process.

\subsection{Our Position}
\begin{wrapfigure}{r}{5cm}
\vspace{-12pt}
    \includegraphics[width=1\linewidth]{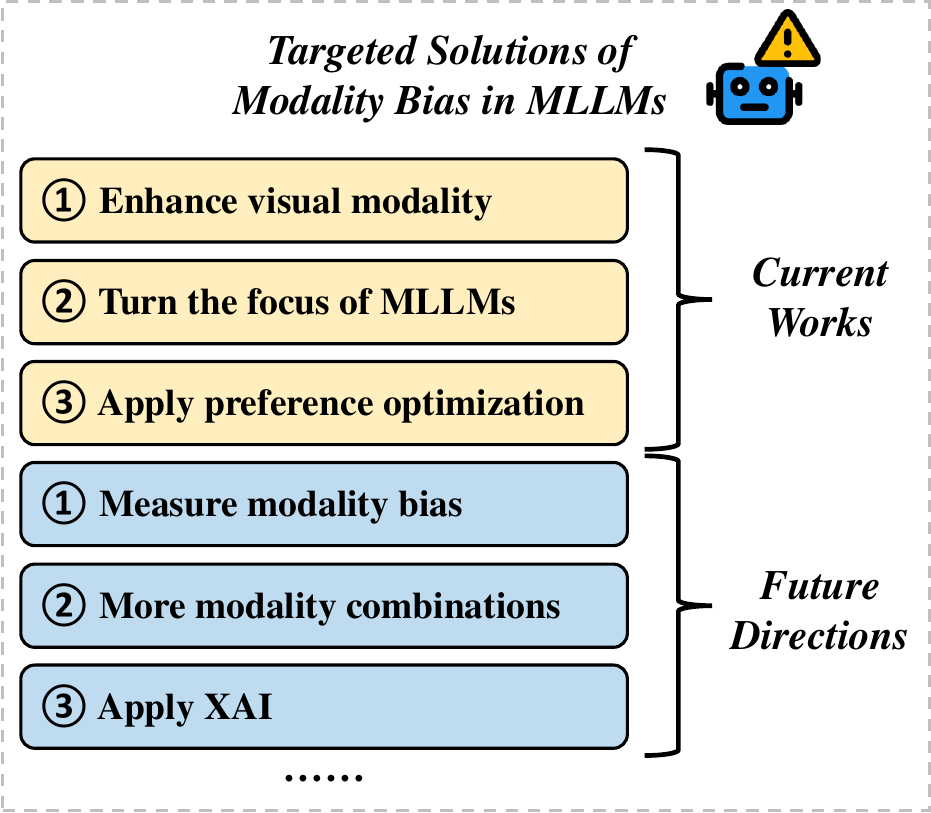}
    \caption{Targeted solutions of modality bias in MLLMs, including current works and future directions.}
    \label{fig:target}
\end{wrapfigure}

In the context of Multimodal Large Language Models (MLLMs), the presence of modality bias is also evident. For instance, empirical results in \cite{zhang2024debiasing} reveal that MLLMs exhibit modality bias in the generated content. Specifically, the output of MLLMs is often primarily influenced by the language model's prior knowledge, rather than by the input images. That is, MLLMs frequently produce confident responses even when relevant images are absent or when incongruent visual input is provided. This phenomenon is also validated by our experiments in Sec.~\ref{case}. Moreover, work \cite{leng2024curse} further confirms that the modality bias in MLLMs stems from the complex interactions between multiple modalities, which complicates the multimodal debiasing process.
Thus, we propose the position that \textbf{MLLMs are deeply affected by modality bias}. Firstly, we offer the definition of modality bias in Sec.~\ref{defi}. Secondly, we review the research roadmap about modality bias in MLLMs in Sec.~\ref{Roadmap}. Thirdly, we analysis the key factors of modality bias in MLLMs in Sec.~\ref{fac}, accompanied with a case study in Sec.~\ref{case}. Finally, we conclude further the targeted solutions of modality bias in MLLMs, including current works and future directions in Sec.~\ref{target} and Fig.~\ref{fig:target}. The big picture of our position is shown in Fig.~\ref{fig:big}.

\section{Proposed Definition of Modality Bias}
\label{defi}
\begin{figure}[h]
    \centering
    \includegraphics[width=1\linewidth]{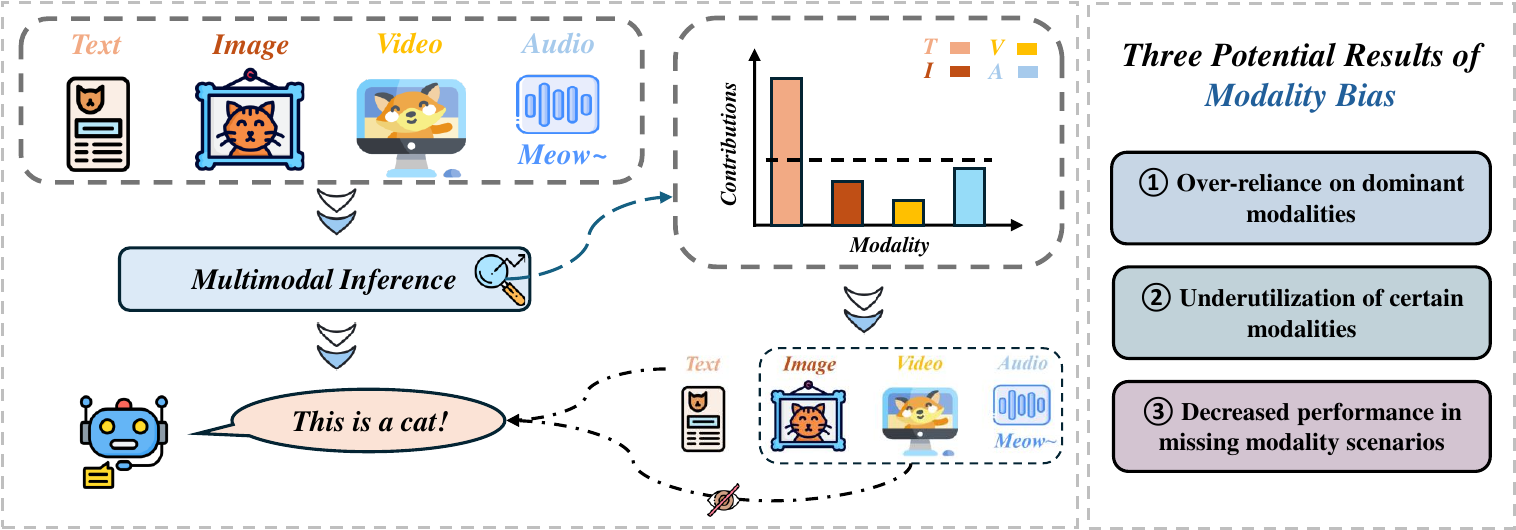}
    \caption{Further definition of modality bias and three potential results.}
    \label{fig:definition}
\end{figure}
Modality bias arises when certain modalities dominate the learning process, while others are underutilized or contribute less effectively~\cite{guo2023modality}. This imbalance often results in a model that is biased towards the dominant modality, thus failing to fully leverage the potential of the underrepresented modalities~\cite{vosoughi2024cross,niu2021counterfactual,gat2020removing,ramakrishnan2018overcoming}. As a result, the model's performance can degrade significantly when the dominant modality is missing, unavailable, or unreliable.


To mathematically describe this imbalance, let us define the contribution of each modality \( M_i \) as \( C(M_i) \), where \( i \in \{1, 2, \dots, n\} \) represents the different modalities (e.g., image, text, audio). The total contribution of all modalities is given by the sum of these individual contributions:

\begin{equation}
   C_{\text{total}} = \sum_{i=1}^{n} C(M_i).
\end{equation}

If certain modalities dominate, the distribution of contributions becomes imbalanced, such that \( C(M_i) \gg \{ C(M_x), C(M_y), \cdots, C(M_z) \}\) for some \( i \neq \{x,y,\cdots,z\}\), as shown in Fig.~\ref{fig:definition}. This imbalance can lead to several issues:

\textbf{\textcircled{1} Over-reliance on dominant modalities:} The model may become overly sensitive to the dominant modality \( M_i \), resulting in biased predictions that fail to incorporate the full diversity of information from the multimodal data.

\textbf{\textcircled{2} Underutilization of certain modalities:} Modalities that are underrepresented in the training data, such as audio or video, contribute less to the learned representations, leading to models that lack robustness when these modalities are needed.

\textbf{\textcircled{3} Decreased performance in missing modality scenarios:} When a dominant modality is missing during inference (for example, if an image is unavailable), the model's performance can drastically drop, as it has not sufficiently learned how to balance the different modalities during training.

To capture the extent of modality bias, we can define a relative measure of imbalance, known as the modality imbalance ratio \( \Delta_{\text{modality}} \), as the ratio of the contribution of the dominant modality to the underutilized modality\footnote{The definition is for better illustration of modality bias, not for calculating.}:

\begin{equation}
\Delta_{\text{modality}} = \frac{C(M_{\text{dominant}})}{C(M_{\text{underutilized}})}.
\end{equation}

This ratio quantifies the disparity between the contributions of the modalities and can serve as a diagnostic tool to identify and address modality bias. A high value of \( \Delta_{\text{modality}} \) indicates a strong bias towards the dominant modality, which can hinder the model's ability to generalize effectively.

In conclusion, modality bias is a fundamental issue in multimodal learning that arises from the unequal contributions of different modalities. It leads to suboptimal learning outcomes and impairs the model's ability to generalize, especially when certain modalities are missing or unavailable. Addressing modality bias involves ensuring that all modalities are effectively utilized and contribute in a balanced manner, thereby improving the robustness and performance of multimodal systems.

\section{How are MLLMs Deeply Affected by Modality Bias?}

\subsection{Research Road-map}

\label{Roadmap}

The exploration process of modality bias in MLLMs can be divided into three directions: \textit{(a) How to prove the bias? (b) How to solve the bias through datasets? (c) How to solve the bias through methods?} These three directions are defined by their different focuses, including bias/debias, datasets/methods, collaborating to highlight and solve modality bias in MLLMs.

\textbf{\textcircled{1} How to prove the bias?} 

With the modality bias in MLLMs emerging gradually as a research focus, several datasets and benchmarks have been proposed to measure the modality bias in MLLMs~\cite{park2025assessing, tong2024eyes,lee2024vlind,leng2024curse,liu2024insight}. 
Park \etal~\cite{park2025assessing} directly proposed a metric named Modality Importance Score (MIS) to measure each modality's contribution in the video question answering task. Based on a comprehensive benchmark, the modality imbalance in current multimodal datasets is proven. 
Lee \etal~\cite{lee2024vlind} and Leng \etal~\cite{leng2024curse} mainly laid emphasis on the modality prior, which is a key reason for modality bias in MLLMs. Specifically, Lee \etal~\cite{lee2024vlind} introduced counterfactual images in VLind-Bench to measure the language priors of LVLMs, proving that LVLMs have a great over-reliance on language priors. Leng \etal~\cite{leng2024curse} proposed a more comprehensive benchmark, namely Curse of Multi-Modalities (CMM), including three modalities: language, visual, and audio. The results of CMM further explain the contributors to hallucinations, where the over-reliance on unimodal priors plays an important role. Liu~\etal~\cite{liu2024insight} explored the bias from the perspective of vision-knowledge conflicts, proving the over-reliance of MLLMs on text queries. Moreover, Tong \etal~\cite{tong2024eyes} proposed the Multimodal Visual Patterns (MMVP) benchmark, further exploring the contrastive language-image pre-training (CLIP)'s weaknesses, which lead to MLLMs' failures in understanding visual information. 

\textbf{\textcircled{2} How to solve the bias through datasets?}

With the modality bias proven to be a common phenomenon in datasets, which is the foundation of training and validation for MLLMs, researchers set their sights on how to solve the bias in datasets~\cite{chen2024we,chen2024quantifying,yue2024mmmu}. Chen~\etal~\cite{chen2024quantifying} proposed MORE, a VQA dataset that requires multi-hop reasoning and overcoming unimodal biases, providing counterexample data to drive the LVLMs to overcome modality bias. Meanwhile, several works focus on decreasing the modality bias in multimodal datasets. Chen~\etal~\cite{chen2024we} proposed MMStar, a meticulously designed multimodal benchmark, of which each sample shows visual dependency, avoiding the modality bias in datasets. Yue~\etal~\cite{yue2024mmmu} built a robust benchmark MMMU-Pro based on MMMU~\cite{yue2024mmmuv1}. Through steps such as making questions embedded in images, MMMU-Pro is equipped with the ability to force MLLMs to both "see" and "read". 

\textbf{\textcircled{3} How to solve the bias through methods?} 

Besides datasets, applying specific methods to reduce the modality bias in MLLMs is another tendency~\cite{zhang2024debiasing,tong2024eyes,zhao2024mmicl,pi2024strengthening,liu2024insight,liu2024paying,zhao2024looking,zhang2025debiasing,li2025devil}. Typically, Pi~\etal~\cite{pi2024strengthening} and Zhang~\etal~\cite{zhang2025debiasing} introduced preference learning methods, such as Bootstrapped Preference Optimization (BPO) and Noise-Aware Preference Optimization (NaPO), solving the modality bias problem based on building negative response datasets. Meanwhile, Zhang~\etal~\cite{zhang2024debiasing}, Liu~\etal~\cite{liu2024insight}, and Tong~\etal~\cite{tong2024eyes} proposed frameworks and methods to "force" MLLMs to pay more attention to images and boost MLLMs' visual understanding abilities. Moreover, Li~\etal~\cite{li2025devil} focused on the Multimodal Reward Models (MM-RMs) for MLLMs, proposing a shortcut-aware MM-RM learning algorithm, decreasing MLLMs' reliance on unimodal spurious correlations. Most above works consider unimodal dependency, especially on textual modality, as the key reason for modality bias. Thus, the boosting of the visual modality gradually turns into a major research topic.

\subsection{Key Factors of Modality Bias in MLLMs}
\label{fac}
Based on Sec~\ref{Roadmap}, the key factors of modality bias in MLLMs can be concluded as follows: dataset imbalances, varying modal capabilities, training objectives, and the interactions between modalities. These factors contribute to the unequal utilization of modalities during training, leading to biases towards certain modalities and suboptimal performance. We summarize three key factors as follows:

\noindent \textbf{\textcircled{1} Dataset Imbalances:} The training dataset composition significantly influences modality utilization. Datasets often have imbalanced modality distributions, where some modalities, such as text or images, are more abundant or have different information density~\cite{chen2024we,chen2024quantifying,yue2024mmmu}. This imbalance leads to models learning representations biased towards the more frequent modalities, while under-utilizing the less represented ones, even when multiple modalities are available. In addition, textual data is often more semantically dense or informative than visual data in certain tasks, due to its structured and explicit nature. As a result, models tend to prioritize textual inputs during learning, treating accompanying modalities such as images merely as auxiliary cues, further amplifying the reliance on dominant modalities.

\noindent \textbf{\textcircled{2} Asymmetric Modal Backbone Capabilities:} Different modalities vary in complexity and in the architectural designs used to process them. Language models often benefit from mature and highly optimized transformer-based architectures~\cite{liu2024deepseek,bi2024deepseek,naveed2023comprehensive}, which are not only effective but also backed by extensive research and industrial-scale pretraining. In contrast, processing visual or acoustic data typically requires more diverse and specialized backbones~\cite{ren2023masked,han2022survey,ren2024bringing,liu2021swin,ren2024sharing,huang2024learning} and may not benefit from equally massive pretraining corpora. Moreover, the rapid advancement of language models, fueled by large-scale datasets and sustained community focus, has further widened the performance gap across modalities. As a result, multimodal models with strong language backbones tend to over-rely on text inputs, under-utilizing other modalities, particularly those that demand more complex or less mature processing pipelines.

\noindent \textbf{\textcircled{3} Training Objectives:} The choice of training objectives fundamentally shapes how multimodal models utilize different modalities and often exacerbates modality bias. Pretraining strategies in many state-of-the-art multimodal models—such as CLIP-style contrastive learning, image–text matching (ITM), masked language modeling (MLM), or caption generation—tend to prioritize text–image alignment due to the abundance of paired data and the relative ease of textual supervision. These objectives implicitly encourage the model to rely heavily on language as the semantic anchor, such as LanguageBind~\cite{languagebind2024} and UniBind~\cite{UniBind2024}. Consequently, modalities like audio, video, point clouds, or thermal data—which are harder to align, less semantically rich in isolation, or lack large-scale supervision—are under-optimized during pretraining. Furthermore, most objectives do not explicitly encourage consistent cross-modal alignment or robust fusion across diverse modalities, resulting in imbalanced feature representations and limited generalization to underrepresented input types.

Additionally, two other factors contribute to modality bias:

\noindent \textbf{\textcircled{4} Differences in Convergence Rates:} Each modality converges at different rates during training. Some modalities, like images and text, are more easily aligned with target labels due to their structure and high information density, while others, such as audio or video, require more complex processing. This disparity results in certain modalities being more influential in the model’s final learned representation, amplifying modality bias.

\noindent \textbf{\textcircled{5} Modal Interactions and Integrations:} The interaction between modalities also affects modality bias. If the relationships between modalities are not explicitly learned, the model may favor the more easily processed modality, like text, over others. The complexity of integrating multimodal information can exacerbate bias, as the model may struggle to effectively combine all modalities, resulting in predictions that under-utilize available data.

In summary, modality bias is driven by factors such as dataset imbalances, differences in modal capabilities, training objectives, and the interactions between modalities. Addressing these factors is essential to mitigating modality bias and improving multimodal model performance. Strategies to balance modality contributions during training, optimize multimodal integration, and address dataset imbalances are critical for building fairer and more robust multimodal systems.

\section{Case Study}
\label{case}

\begin{figure}[h]
    \centering
    \includegraphics[width=1\linewidth]{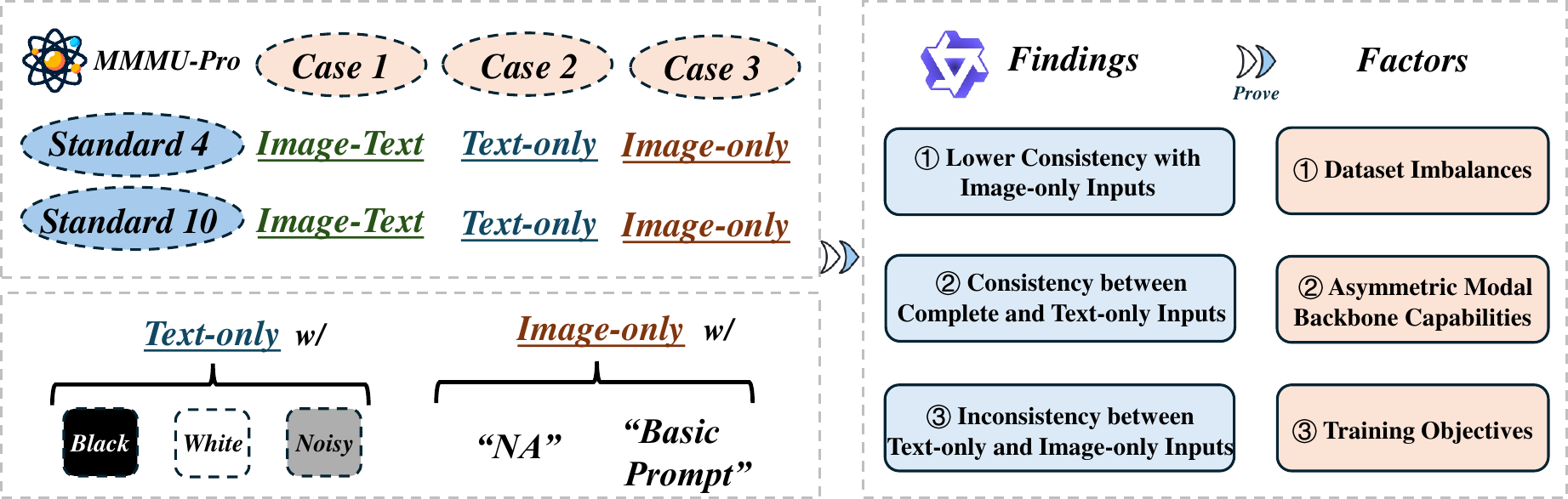}
    \caption{Case study for exploring modality bias in MLLMs. Dataset: MMMU-Pro, MLLM: Qwen2.5VL. Based on this case study, the three main factors proposed in Sec.~\ref{fac} are further illustrated and proved. "white" means the image pixels are all set to 255. "black" means the image pixels are all set to 0.}
    \label{fig:enter-label}
\end{figure}

\begin{table*}[h!]
\centering
\renewcommand{\tabcolsep}{4pt}
\caption{Directly applying missing modality evaluation with MVLMs on MMMU-pro dataset. Basic prompt (B-P) for Image only is:  "Based on the provided images, please answer the question."}
\label{exp}
\resizebox{0.85\linewidth}{!}{
\begin{tabular}{c|c|ccc|ccc}
\toprule
\multirow{2}{*}{Model}              & \multirow{2}{*}{Inference} & \multicolumn{3}{c|}{Standard 4}                                     & \multicolumn{3}{c}{Standard 10}                                    \\ \cmidrule{3-8} 
                                    &                            & \multicolumn{1}{c|}{Image-Text} & \multicolumn{1}{c|}{Text /w white} & Image & \multicolumn{1}{c|}{Image-Text} & \multicolumn{1}{c|}{Text w/ white} & Image w/ B-P \\ \midrule
\multirow{5}{*}{Qwen2.5-VL-7B-I} & \multirow{2}{*}{Direct}  & 48.32 & 34.91 & 28.73  & 37.57 & 21.73 & 14.91 \\ \cmidrule{3-8} 
& & - & \cellcolor{red!13}-13.41 $\downarrow$ & \cellcolor{red!19}-19.59 $\downarrow$ & - & \cellcolor{red!15}-15.84 $\downarrow$ & \cellcolor{red!22}-22.66 $\downarrow$   \\ \cmidrule{2-8} 
& \multirow{2}{*}{CoT}   & 52.14 & 35.20 & 28.38 & 39.94 & 21.39 & 14.97 \\ \cmidrule{3-8} 
 & & - & \cellcolor{red!16}-16.94 $\downarrow$ & \cellcolor{red!23}-23.76 $\downarrow$ & - & \cellcolor{red!18}-18.55 $\downarrow$ & \cellcolor{red!24}-24.97 $\downarrow$ \\ \midrule
 \multirow{5}{*}{Qwen2.5-VL-32B-I} & \multirow{2}{*}{Direct} & 57.80 & 40.17 & 27.23 & 43.94 & 28.32 & 15.38 \\ \cmidrule{3-8} 
& & - &  \cellcolor{red!17}-17.63 $\downarrow$ & \cellcolor{red!30}-30.57 $\downarrow$& - & \cellcolor{red!15}-15.62 $\downarrow$  & \cellcolor{red!28}-28.56 $\downarrow$  \\ \cmidrule{2-8} 
& \multirow{2}{*}{CoT} & 59.88 & 40.12 & 26.36 & 48.55 & 26.82 & 14.34 \\ \cmidrule{3-8} 
 & & - & \cellcolor{red!19}-19.76 $\downarrow$ & \cellcolor{red!33}-33.52 $\downarrow$ & - & \cellcolor{red!21}-21.73 $\downarrow$ & \cellcolor{red!34}-34.21 $\downarrow$  \\ \midrule
\multirow{2}{*}{Model}              & \multirow{2}{*}{Inference} & \multicolumn{3}{c|}{Standard 4}                                     & \multicolumn{3}{c}{Standard 10}                                    \\ \cmidrule{3-8} 
                                    &                            & \multicolumn{1}{c|}{Image-Text} & \multicolumn{1}{c|}{Text /w black} & Image & \multicolumn{1}{c|}{Image-Text} & \multicolumn{1}{c|}{Text w/ black} & Image w/ B-P \\ \midrule
\multirow{5}{*}{Qwen2.5-VL-7B-I} & \multirow{2}{*}{Direct}  & 48.32 & 34.86 & 29.02 & 37.63 & 21.62 & 14.51 \\ \cmidrule{3-8} 
& & - & \cellcolor{red!13}-13.46 $\downarrow$ & \cellcolor{red!19}-19.30 $\downarrow$ & - & \cellcolor{red!16}-16.01 $\downarrow$ & \cellcolor{red!23}-23.12 $\downarrow$ \\ \cmidrule{2-8} 
& \multirow{2}{*}{CoT}   & 52.08 & 21.45 & 28.38 & 39.94 & 21.45 & 14.86 \\ \cmidrule{3-8} 
 & & - & \cellcolor{red!30}-30.63 $\downarrow$ & \cellcolor{red!23}-23.70 $\downarrow$ & - & \cellcolor{red!18}-18.49 $\downarrow$ & \cellcolor{red!25}-25.08 $\downarrow$ \\ \midrule
 \multirow{5}{*}{Qwen2.5-VL-32B-I} & \multirow{2}{*}{Direct} & 57.80 & 40.17 & 26.99 & 43.87 & 28.21 & 15.03 \\ \cmidrule{3-8} 
& & - & \cellcolor{red!17}-17.63 $\downarrow$ & \cellcolor{red!30}-30.81 $\downarrow$ & - & \cellcolor{red!15}-15.66 $\downarrow$ & \cellcolor{red!28}-28.84 $\downarrow$ \\ \cmidrule{2-8} 
& \multirow{2}{*}{CoT} & 59.83 & 40.35 & 26.65 & 48.44 & 26.82 & 14.86 \\ \cmidrule{3-8} 
 & & - & \cellcolor{red!19}-19.48 $\downarrow$ & \cellcolor{red!33}-33.18 $\downarrow$ & - & \cellcolor{red!21}-21.62 $\downarrow$ & \cellcolor{red!33}-33.58 $\downarrow$ \\ \midrule
\multirow{2}{*}{Model}              & \multirow{2}{*}{Inference} & \multicolumn{3}{c|}{Standard 4}                                     & \multicolumn{3}{c}{Standard 10}                                    \\ \cmidrule{3-8} 
                                    &                            & \multicolumn{1}{c|}{Image-Text} & \multicolumn{1}{c|}{Text w/ noise} & Image & \multicolumn{1}{c|}{Image-Text} & \multicolumn{1}{c|}{Text w/ noise} & Image w/ B-P \\ \midrule
\multirow{5}{*}{Qwen2.5-VL-7B-I} & \multirow{2}{*}{Direct}  & 48.32 & 34.97 & 26.99 & 37.57 & 21.56 & 14.86 \\ \cmidrule{3-8} 
                                       &                            & - & \cellcolor{red!13}-13.35 $\downarrow$ & \cellcolor{red!21}-21.33 $\downarrow$ & - & \cellcolor{red!16}-16.01 $\downarrow$ & \cellcolor{red!22}-22.71 $\downarrow$ \\ \cmidrule{2-8} 
                                       & \multirow{2}{*}{CoT}      & 51.73 & 35.66 & 27.40 & 40.00 & 21.27 & 13.35  \\ \cmidrule{3-8} 
                                       &                            & - & \cellcolor{red!16}-16.07 $\downarrow$ & \cellcolor{red!24}-24.33 $\downarrow$ & - & \cellcolor{red!18}-18.73 $\downarrow$ & \cellcolor{red!26}-26.65 $\downarrow$ \\ \midrule
\multirow{5}{*}{Qwen2.5-VL-32B-I} & \multirow{2}{*}{Direct} & 57.75 & 40.23 & 26.13 & 43.70 & 28.27 & 14.28 \\ \cmidrule{3-8} 
                                       &                            & - & \cellcolor{red!17}-17.52 $\downarrow$ & \cellcolor{red!31}-31.62 $\downarrow$ & - & \cellcolor{red!15}-15.43 $\downarrow$ & \cellcolor{red!29}-29.42 $\downarrow$ \\ \cmidrule{2-8} 
                                       & \multirow{2}{*}{CoT}     & 60.12 & 40.12 & 28.03 & 48.03 & 27.17 & 13.99 \\ \cmidrule{3-8} 
                                       &                           & - & \cellcolor{red!20}-20.00 $\downarrow$ & \cellcolor{red!32}-32.09 $\downarrow$ & - & \cellcolor{red!20}-20.86 $\downarrow$ & \cellcolor{red!34}-34.04 $\downarrow$ \\ \midrule
\end{tabular}}
\vspace{-12pt}
\end{table*}

The results presented in Tab.~\ref{exp} and Tab.~\ref{tab:consistency_analysis} reveal several key insights regarding the performance of the multimodal large model when tested with different input combinations across the MMMU-pro dataset. The process of case study is shown in Fig.~\ref{case} These insights can be linked to the three key factors identified in our analysis of modality bias in Multimodal Large Language Models (MLLMs): \textcircled{1} dataset imbalances, \textcircled{2} asymmetric modal backbone capabilities, and \textcircled{3} training objectives.

\noindent \textbf{Lower Consistency with Image-only Inputs} 27.17\% (Complete \& Image-only, Direct) and 28.21\% (Complete \& Image-only, CoT): The relatively low consistency between the complete input and image-only input suggests that the image modality alone is not sufficient for the model to make consistent predictions. When the model only has access to visual data, its predictions tend to be less reliable, underscoring the inadequacies of the model in processing visual data in isolation. This result supports the factor of \textcircled{1} dataset imbalances, where the richness and complexity of image data, compared to more compact textual data, pose challenges for the model. Although images provide important visual cues, the model struggles to effectively utilize the image modality alone, indicating that the image modality is underutilized in the absence of complementary text data.

\noindent \textbf{Consistency between Complete and Text-only Inputs} 56.53\% (Complete \& Text-only, Direct) and 43.64\% (Complete \& Text-only, CoT): The finding that over half of the samples show consistency between the complete (both image and text) and text-only inputs across both inference techniques (Direct and CoT) is significant. It suggests that textual information alone is a strong foundation for the model's predictions, and in many cases, the image modality does not substantially alter the model’s output. This highlights the dominance of the language modality, which is particularly advantageous due to its well-established processing capabilities. This result is consistent with the factor of \textcircled{2} asymmetric modal backbone capabilities, where models with stronger language backbones, such as this one, tend to perform better on language tasks, often overshadowing the visual modality and limiting the model's ability to effectively integrate multimodal information.

\noindent \textbf{Inconsistency between Text-only and Image-only Inputs} 26.76\% (Text-only \& Image-only, Direct) and 25.95\% (Text-only \& Image-only, CoT): The low consistency between text-only and image-only inputs highlights the challenge the model faces when dealing with these two distinct modalities separately. This discrepancy suggests that both text and image provide complementary yet crucial information for accurate predictions. Textual data offers rich semantic context, nuances, and detail that images alone cannot convey, while images provide visual cues and spatial relationships that text cannot fully express. The low consistency between these two modalities, especially in the CoT setting, where reasoning and integration are critical, points to the challenge of combining these modalities effectively. This underscores the factor of \textcircled{3} training objectives, where existing training strategies often fail to adequately balance multimodal learning, leading to modality-specific shortcuts. In the case of this model, the failure to effectively combine text and image information results in inconsistent predictions, especially when reasoning across modalities is required.


Our findings underscore the importance of balanced training strategies and model architectures to address modality bias and improve multimodal integration. This also highlights the need for future research aimed at developing MLLMs that can more effectively process and combine diverse sources of information, mitigating the impact of modality bias.

\begin{table*}[t!]
\centering
\caption{Prediction consistency analysis of Qwen2.5-VL-7B-Instruct model on MMMU-pro. Complete means both images and text are used as input; Direct and CoT refer to the inference techniques.}
\renewcommand{\tabcolsep}{8pt}
\label{tab:consistency_analysis}
\resizebox{0.9\linewidth}{!}{
\begin{tabular}{ll|cc|cc|cc|cc}
\toprule
\multirow{3}{*}{Choices} & & \multicolumn{2}{c|}{Standard 4 (Direct)} & \multicolumn{2}{c|}{Standard 4 (CoT)} & \multicolumn{2}{c|}{Standard 10 (Direct)} & \multicolumn{2}{c}{Standard 10 (CoT)} \\ \cmidrule{3-10}
& & Num. & Percentage & Num. & Percentage & Num. & Percentage & Num. & Percentage \\ \midrule
\multicolumn{10}{c}{All Samples} \\ \midrule
\multirow{4}{*}{Consistent} & Complete \& Text-only   & 978  & 56.53\% & 755  & 43.64\% & 821 & 47.46\% & 568 & 32.83\% \\ \cmidrule{2-10}
& Complete \& Image-only  & 470  & 27.17\% & 488  & 28.21\% & 262 & 15.14\% & 260 & 15.03\% \\ \cmidrule{2-10}
& Text-only \& Image-only & 463  & 26.76\% & 449  & 25.95\% & 234 & 13.53\% & 232 & 13.41\% \\ \cmidrule{2-10}
& All                     & 267  & 15.43\% & 220  & 12.72\% & 112 & 6.47\% & 87 & 5.03\% \\ \midrule
\multirow{4}{*}{Inconsistent} & Complete \& Text-only   & 752  & 43.47\% & 975  & 56.36\% & 909 & 52.54\% & 1162 & 67.17\% \\ \cmidrule{2-10}
& Complete \& Image-only  & 1260 & 72.83\% & 1242 & 71.70\% & 1468 & 84.86\% & 1470 & 84.97\% \\ \cmidrule{2-10}
& Text-only \& Image-only & 1267 & 73.24\% & 1281 & 74.05\% & 1496 & 86.47\% & 1498 & 86.59\% \\ \cmidrule{2-10}
& All                     & 353  & 20.40\% & 478  & 27.63\% & 637 & 36.82\% & 844 & 48.79\% \\ \midrule
\multicolumn{10}{c}{Correct Samples} \\ \midrule
\multirow{4}{*}{Consistent} & Complete \& Text-only   & 456 & 26.32\% & 417 & 24.08\% & 284 & 16.42\% & 231 & 13.36\% \\ \cmidrule{2-10}
 & Complete \& Image-only  & 254 & 14.67\% & 264 & 15.26\% & 126 & 7.28\% & 135 & 7.80\% \\ \cmidrule{2-10}
& Text-only \& Image-only & 185 & 10.68\% & 174 & 10.05\% & 61 & 3.53\% & 56 & 3.24\% \\ \cmidrule{2-10}
& All                     & 142 & 8.20\%  & 125 & 7.22\%  & 49  & 2.83\%  & 42  & 2.43\% \\ \midrule
\multicolumn{10}{c}{Wrong Samples} \\ \midrule
\multirow{4}{*}{Consistent} & Complete \& Text-only   & 522 & 30.15\% & 338 & 19.51\% & 537 & 31.04\% & 337 & 19.46\% \\ \cmidrule{2-10}
& Complete \& Image-only  & 216 & 12.49\% & 224 & 12.94\% & 136 & 7.86\% & 125 & 7.22\% \\ \cmidrule{2-10}
& Text-only \& Image-only & 278 & 16.06\% & 275 & 15.91\% & 173 & 9.99\% & 176 & 10.16\% \\ \cmidrule{2-10}
 & All                     & 125 & 7.22\%  & 95  & 5.49\%  & 63  & 3.64\%  & 45  & 2.60\% \\ \midrule
\end{tabular}}
\end{table*}

\section{Targeted Solutions}
\label{target}
\subsection{Current Works}

\textbf{\textcircled{1} Enhance visual modality's contribution in datasets:} With the in-depth exploration in modality bias, especially in the vision-language modality combination, visual information tends to be proven to be ignored, resulting in MLLMs' over-reliance on the textual modality~\cite{zhang2024debiasing}. Thus, researchers naturally attempt to enhance visual modality's contribution in datasets to balance the information from different modalities. Typical cases include MMStar~\cite{chen2024we} and MMMU-Pro~\cite{yue2024mmmu}, where MMStar carefully selects visually dependent samples and MMMU-Pro not only filters out visually independent samples but also embeds questions into images. \textit{Such works provide an optimization direction for current multimodal datasets. While few works contain a systematic index to evaluate the modality bias in datasets~\cite{park2025assessing}, the others tend to prove the datasets' necessity based on MLLMs' disappointing results. More modality bias evaluation methods need to be explored to construct a better debias multimodal datasets.} Importantly, the feedback from MLLMs on these datasets should also be considered, as their performance can inform how datasets are optimized, offering valuable insights for future improvements.

\textbf{\textcircled{2} Turn the focus of MLLMs from textual information into visual information:} Considering the ignorance 
 of visual modality in the inference of MLLMs, it is an intuitive approach to force MLLMs to lay more emphasis on visual modality. 
 Works such as~\cite{liu2024paying, zhang2024debiasing} apply strategies, 
 mostly training-free, to guide MLLMs towards visual modality.
 While Zhao~\etal\cite{zhao2024looking} proposed a novel framework to help MLLMs compress the influence of textual bias, enhancing visual modality across the model. \textit{However, due to the excessive focus on the visual modality in such works, there's difficulty in applying them to broader modality bias situations. The real-world application requires the exploration of bias in richer modality combination.}

\textbf{\textcircled{3} Apply preference optimization strategies:} Besides the adjustments of multimodal datasets' content and MLLMs' focus, another popular method is to use a preference optimization strategy to correct modality bias internally~\cite{pi2024strengthening,zhang2025debiasing}. Pi~\etal~\cite{pi2024strengthening} built a preference dataset containing samples reflecting modality bias generated from the pretraining process. Zhang~\etal~\cite{zhang2025debiasing} forced MLLMs to generate answers according to a specific modality through adding noise, thus creating the preference dataset. \textit{Considering solving modality bias as a preference optimization goal is a creative and reasonable idea that brings new insights to researchers. However, existing preference optimization methods only generate modality bias samples in limited ways, while the real causes of modality bias are complex and multi-stage, which await further exploration.}

\subsection{Future Directions}

\textbf{\textcircled{1} Measure modality bias in MLLMs:} The exploration of an objective and systematic metric to measure modality bias is crucial for the development of related research. For example, for dataset construction, a metric is needed as a flag that offers researchers a clear direction to make progress.  Fields like semantic segmentation~\cite{rahman2016optimizing,rezatofighi2019generalized} and image restoration~\cite{hu2020subjective} have seen a huge development with the existence and optimization of evaluation metrics, where modality bias in MLLMs still remains almost blank. Therefore, more research works are being called for regarding measuring modality bias in MLLMs.

\textbf{\textcircled{2} Explore modality bias in more modality combinations:} Despite several works attempting to address the modality bias problem, the research focus is mainly set on the modality bias in LVLMs, which is part of the MLLMs. Although the textual information and visual information show great importance for world understanding~\cite{xu2024lvlm}, modalities like audio and tactile also matter~\cite{liu2024valor,dave2024multimodal,lyu2024omnibind}. As to the robotics field~\cite{zhang2025soft,kirschner2025categorizing,agarwal2025vision}, tactile information is indispensable for robots to understand environments and handle downstream tasks such as dexterous manipulation~\cite{gbagbe2024bi}. Due to the modality limitation of current debiasing methods, it is hard for them to be applied in broader situations, hindering their applications in the real world. Thus, more generalized debiasing strategies are required in real-world applications to handle conditions that are more complex and have more modalities besides images and texts.

\textbf{\textcircled{3} Apply XAI for modality bias in MLLMs:} Last but not least, finding the causes of modality bias in MLLMs and visualizing them will have a great positive influence on future works. Even though current works attempt to dig out the reasons for modality bias in MLLMs, they propose opinions from the phenomenon level. The internal mechanism of modality bias still lacks exploration, which is theoretical evidence and guidance to support future works. Thus, explainable AI~\cite{bennetot2024practical,dwivedi2023explainable} is required here, such as visualizing the interaction process between modalities, to deeply analyze the theoretical causes and working mechanism of modality bias in MLLMs, which can be a more solid inspiration for future works.

\section{Conclusion}
This paper aims to highlight the phenomenon of modality bias in MLLMs and call for research work targeted at better integrating multiple modalities. Our position is that \textbf{MLLMs are deeply affected by modality bias}, which is proved and explored by both the theoretical analysis and case study. Moreover, we offer an in-depth discussion targeted at modality bias in MLLMs, including the key factors, potential results, and targeted solutions, hoping to bring new insights to the development of more robust and generalizable multimodal systems.


\bibliographystyle{ieeetr}  
\bibliography{ref}

\begin{thebibliography}{10}

\bibitem{bai2025qwen2}
S.~Bai, K.~Chen, X.~Liu, J.~Wang, W.~Ge, S.~Song, K.~Dang, P.~Wang, S.~Wang, J.~Tang, {\em et~al.}, ``Qwen2. 5-vl technical report,'' {\em arXiv preprint arXiv:2502.13923}, 2025.

\bibitem{wang2024qwen2}
P.~Wang, S.~Bai, S.~Tan, S.~Wang, Z.~Fan, J.~Bai, K.~Chen, X.~Liu, J.~Wang, W.~Ge, {\em et~al.}, ``Qwen2-vl: Enhancing vision-language model's perception of the world at any resolution,'' {\em arXiv preprint arXiv:2409.12191}, 2024.

\bibitem{zhu2025internvl3}
J.~Zhu, W.~Wang, Z.~Chen, Z.~Liu, S.~Ye, L.~Gu, Y.~Duan, H.~Tian, W.~Su, J.~Shao, {\em et~al.}, ``Internvl3: Exploring advanced training and test-time recipes for open-source multimodal models,'' {\em arXiv preprint arXiv:2504.10479}, 2025.

\bibitem{chen2024expanding}
Z.~Chen, W.~Wang, Y.~Cao, Y.~Liu, Z.~Gao, E.~Cui, J.~Zhu, S.~Ye, H.~Tian, Z.~Liu, {\em et~al.}, ``Expanding performance boundaries of open-source multimodal models with model, data, and test-time scaling,'' {\em arXiv preprint arXiv:2412.05271}, 2024.

\bibitem{team2024gemini}
P.~Georgiev, V.~I. Lei, R.~Burnell, L.~Bai, A.~Gulati, G.~Tanzer, D.~Vincent, Z.~Pan, S.~Wang, {\em et~al.}, ``Gemini 1.5: Unlocking multimodal understanding across millions of tokens of context,'' {\em arXiv preprint arXiv:2403.05530}, 2024.

\bibitem{hurst2024gpt}
A.~Hurst, A.~Lerer, A.~P. Goucher, A.~Perelman, A.~Ramesh, A.~Clark, A.~Ostrow, A.~Welihinda, A.~Hayes, A.~Radford, {\em et~al.}, ``Gpt-4o system card,'' {\em arXiv preprint arXiv:2410.21276}, 2024.

\bibitem{dahiya2009tactile}
R.~S. Dahiya, G.~Metta, M.~Valle, and G.~Sandini, ``Tactile sensing—from humans to humanoids,'' {\em IEEE transactions on robotics}, vol.~26, no.~1, pp.~1--20, 2009.

\bibitem{zou2017novel}
L.~Zou, C.~Ge, Z.~J. Wang, E.~Cretu, and X.~Li, ``Novel tactile sensor technology and smart tactile sensing systems: A review,'' {\em Sensors}, vol.~17, no.~11, p.~2653, 2017.

\bibitem{chi2018recent}
C.~Chi, X.~Sun, N.~Xue, T.~Li, and C.~Liu, ``Recent progress in technologies for tactile sensors,'' {\em Sensors}, vol.~18, no.~4, p.~948, 2018.

\bibitem{gallego2020event}
G.~Gallego, T.~Delbr{\"u}ck, G.~Orchard, C.~Bartolozzi, B.~Taba, A.~Censi, S.~Leutenegger, A.~J. Davison, J.~Conradt, K.~Daniilidis, {\em et~al.}, ``Event-based vision: A survey,'' {\em IEEE transactions on pattern analysis and machine intelligence}, vol.~44, no.~1, pp.~154--180, 2020.

\bibitem{zheng2023deep}
X.~Zheng, Y.~Liu, Y.~Lu, T.~Hua, T.~Pan, W.~Zhang, D.~Tao, and L.~Wang, ``Deep learning for event-based vision: A comprehensive survey and benchmarks,'' {\em arXiv preprint arXiv:2302.08890}, 2023.

\bibitem{rebecq2019high}
H.~Rebecq, R.~Ranftl, V.~Koltun, and D.~Scaramuzza, ``High speed and high dynamic range video with an event camera,'' {\em IEEE transactions on pattern analysis and machine intelligence}, vol.~43, no.~6, pp.~1964--1980, 2019.

\bibitem{zheng2024360sfuda++}
X.~Zheng, P.~Y. Zhou, A.~V. Vasilakos, and L.~Wang, ``360sfuda++: Towards source-free uda for panoramic segmentation by learning reliable category prototypes,'' {\em IEEE Transactions on Pattern Analysis and Machine Intelligence}, 2024.

\bibitem{zheng2024semantics}
X.~Zheng, P.~Zhou, A.~V. Vasilakos, and L.~Wang, ``Semantics distortion and style matter: Towards source-free uda for panoramic segmentation,'' in {\em Proceedings of the IEEE/CVF Conference on Computer Vision and Pattern Recognition}, pp.~27885--27895, 2024.

\bibitem{zhong2025omnisam}
D.~Zhong, X.~Zheng, C.~Liao, Y.~Lyu, J.~Chen, S.~Wu, L.~Zhang, and X.~Hu, ``Omnisam: Omnidirectional segment anything model for uda in panoramic semantic segmentation,'' {\em arXiv preprint arXiv:2503.07098}, 2025.

\bibitem{liu2024mmbench}
Y.~Liu, H.~Duan, Y.~Zhang, B.~Li, S.~Zhang, W.~Zhao, Y.~Yuan, J.~Wang, C.~He, Z.~Liu, {\em et~al.}, ``Mmbench: Is your multi-modal model an all-around player?,'' in {\em European conference on computer vision}, pp.~216--233, Springer, 2024.

\bibitem{fang2024mmbench}
X.~Fang, K.~Mao, H.~Duan, X.~Zhao, Y.~Li, D.~Lin, and K.~Chen, ``Mmbench-video: A long-form multi-shot benchmark for holistic video understanding,'' {\em Advances in Neural Information Processing Systems}, vol.~37, pp.~89098--89124, 2024.

\bibitem{mathew2021docvqa}
M.~Mathew, D.~Karatzas, and C.~Jawahar, ``Docvqa: A dataset for vqa on document images,'' in {\em Proceedings of the IEEE/CVF winter conference on applications of computer vision}, pp.~2200--2209, 2021.

\bibitem{li2024survey}
L.~Li, G.~Chen, H.~Shi, J.~Xiao, and L.~Chen, ``A survey on multimodal benchmarks: In the era of large ai models,'' {\em arXiv preprint arXiv:2409.18142}, 2024.

\bibitem{li2024survey1}
J.~Li, W.~Lu, H.~Fei, M.~Luo, M.~Dai, M.~Xia, Y.~Jin, Z.~Gan, D.~Qi, C.~Fu, {\em et~al.}, ``A survey on benchmarks of multimodal large language models,'' {\em arXiv preprint arXiv:2408.08632}, 2024.

\bibitem{wu2024visual}
J.~Wu, Z.~Zhang, Y.~Xia, X.~Li, Z.~Xia, A.~Chang, T.~Yu, S.~Kim, R.~A. Rossi, R.~Zhang, {\em et~al.}, ``Visual prompting in multimodal large language models: A survey,'' {\em arXiv preprint arXiv:2409.15310}, 2024.

\bibitem{jiang2025survey}
C.~Jiang, Z.~Wang, M.~Dong, and J.~Gui, ``Survey of adversarial robustness in multimodal large language models,'' {\em arXiv preprint arXiv:2503.13962}, 2025.

\bibitem{huo2025continue}
Y.~Huo and H.~Tang, ``When continue learning meets multimodal large language model: A survey,'' {\em arXiv preprint arXiv:2503.01887}, 2025.

\bibitem{zhang2024debiasing}
Y.-F. Zhang, W.~Yu, Q.~Wen, X.~Wang, Z.~Zhang, L.~Wang, R.~Jin, and T.~Tan, ``Debiasing multimodal large language models,'' {\em arXiv preprint arXiv:2403.05262}, 2024.

\bibitem{park2025assessing}
J.~Park, K.~J. Jang, B.~Alasaly, S.~Mopidevi, A.~Zolensky, E.~Eaton, I.~Lee, and K.~Johnson, ``Assessing modality bias in video question answering benchmarks with multimodal large language models,'' in {\em Proceedings of the AAAI Conference on Artificial Intelligence}, vol.~39, pp.~19821--19829, 2025.

\bibitem{tong2024eyes}
S.~Tong, Z.~Liu, Y.~Zhai, Y.~Ma, Y.~LeCun, and S.~Xie, ``Eyes wide shut? exploring the visual shortcomings of multimodal llms,'' in {\em Proceedings of the IEEE/CVF Conference on Computer Vision and Pattern Recognition}, pp.~9568--9578, 2024.

\bibitem{chen2024we}
L.~Chen, J.~Li, X.~Dong, P.~Zhang, Y.~Zang, Z.~Chen, H.~Duan, J.~Wang, Y.~Qiao, D.~Lin, {\em et~al.}, ``Are we on the right way for evaluating large vision-language models?,'' {\em arXiv preprint arXiv:2403.20330}, 2024.

\bibitem{chen2024quantifying}
M.~Chen, Y.~Cao, Y.~Zhang, and C.~Lu, ``Quantifying and mitigating unimodal biases in multimodal large language models: A causal perspective,'' {\em arXiv preprint arXiv:2403.18346}, 2024.

\bibitem{yue2024mmmu}
X.~Yue, T.~Zheng, Y.~Ni, Y.~Wang, K.~Zhang, S.~Tong, Y.~Sun, B.~Yu, G.~Zhang, H.~Sun, {\em et~al.}, ``Mmmu-pro: A more robust multi-discipline multimodal understanding benchmark,'' {\em arXiv preprint arXiv:2409.02813}, 2024.

\bibitem{xu2023multimodal}
P.~Xu, X.~Zhu, and D.~A. Clifton, ``Multimodal learning with transformers: A survey,'' {\em IEEE Transactions on Pattern Analysis and Machine Intelligence}, vol.~45, no.~10, pp.~12113--12132, 2023.

\bibitem{bayoudh2022survey}
K.~Bayoudh, R.~Knani, F.~Hamdaoui, and A.~Mtibaa, ``A survey on deep multimodal learning for computer vision: advances, trends, applications, and datasets,'' {\em The Visual Computer}, vol.~38, no.~8, pp.~2939--2970, 2022.

\bibitem{jabeen2023review}
S.~Jabeen, X.~Li, M.~S. Amin, O.~Bourahla, S.~Li, and A.~Jabbar, ``A review on methods and applications in multimodal deep learning,'' {\em ACM Transactions on Multimedia Computing, Communications and Applications}, vol.~19, no.~2s, pp.~1--41, 2023.

\bibitem{zheng2024learning}
X.~Zheng, Y.~Lyu, and L.~Wang, ``Learning modality-agnostic representation for semantic segmentation from any modalities,'' in {\em European Conference on Computer Vision}, pp.~146--165, Springer, 2024.

\bibitem{liao2025memorysam}
C.~Liao, X.~Zheng, Y.~Lyu, H.~Xue, Y.~Cao, J.~Wang, K.~Yang, and X.~Hu, ``Memorysam: Memorize modalities and semantics with segment anything model 2 for multi-modal semantic segmentation,'' {\em arXiv preprint arXiv:2503.06700}, 2025.

\bibitem{brodermann2025cafuser}
T.~Br{\"o}dermann, C.~Sakaridis, Y.~Fu, and L.~Van~Gool, ``Cafuser: Condition-aware multimodal fusion for robust semantic perception of driving scenes,'' {\em IEEE Robotics and Automation Letters}, 2025.

\bibitem{manzoor2023multimodality}
M.~A. Manzoor, S.~Albarri, Z.~Xian, Z.~Meng, P.~Nakov, and S.~Liang, ``Multimodality representation learning: A survey on evolution, pretraining and its applications,'' {\em ACM Transactions on Multimedia Computing, Communications and Applications}, vol.~20, no.~3, pp.~1--34, 2023.

\bibitem{wei2024enhancing}
Y.~Wei, R.~Feng, Z.~Wang, and D.~Hu, ``Enhancing multimodal cooperation via sample-level modality valuation,'' in {\em Proceedings of the IEEE/CVF Conference on Computer Vision and Pattern Recognition}, pp.~27338--27347, 2024.

\bibitem{zheng2024centering}
X.~Zheng, Y.~Lyu, J.~Zhou, and L.~Wang, ``Centering the value of every modality: Towards efficient and resilient modality-agnostic semantic segmentation,'' in {\em European Conference on Computer Vision}, pp.~192--212, Springer, 2024.

\bibitem{zheng2025reducing}
X.~Zheng, Y.~Lyu, L.~Jiang, D.~P. Paudel, L.~Van~Gool, and X.~Hu, ``Reducing unimodal bias in multi-modal semantic segmentation with multi-scale functional entropy regularization,'' {\em arXiv preprint arXiv:2505.06635}, 2025.

\bibitem{zheng2024learning1}
X.~Zheng, H.~Xue, J.~Chen, Y.~Yan, L.~Jiang, Y.~Lyu, K.~Yang, L.~Zhang, and X.~Hu, ``Learning robust anymodal segmentor with unimodal and cross-modal distillation,'' {\em arXiv preprint arXiv:2411.17141}, 2024.

\bibitem{peng2022balanced}
X.~Peng, Y.~Wei, A.~Deng, D.~Wang, and D.~Hu, ``Balanced multimodal learning via on-the-fly gradient modulation,'' in {\em Proceedings of the IEEE/CVF conference on computer vision and pattern recognition}, pp.~8238--8247, 2022.

\bibitem{alabdulmohsinclip}
I.~Alabdulmohsin, X.~Wang, A.~P. Steiner, P.~Goyal, A.~D'Amour, and X.~Zhai, ``Clip the bias: How useful is balancing data in multimodal learning?,'' in {\em The Twelfth International Conference on Learning Representations}.

\bibitem{radford2021learning}
A.~Radford, J.~W. Kim, C.~Hallacy, A.~Ramesh, G.~Goh, S.~Agarwal, G.~Sastry, A.~Askell, P.~Mishkin, J.~Clark, {\em et~al.}, ``Learning transferable visual models from natural language supervision,'' in {\em International conference on machine learning}, pp.~8748--8763, PmLR, 2021.

\bibitem{tschannen2023clippo}
M.~Tschannen, B.~Mustafa, and N.~Houlsby, ``Clippo: Image-and-language understanding from pixels only,'' in {\em Proceedings of the IEEE/CVF Conference on Computer Vision and Pattern Recognition}, pp.~11006--11017, 2023.

\bibitem{yang2024clip}
C.~Yang, Z.~An, L.~Huang, J.~Bi, X.~Yu, H.~Yang, B.~Diao, and Y.~Xu, ``Clip-kd: An empirical study of clip model distillation,'' in {\em Proceedings of the IEEE/CVF Conference on Computer Vision and Pattern Recognition}, pp.~15952--15962, 2024.

\bibitem{wu2023tinyclip}
K.~Wu, H.~Peng, Z.~Zhou, B.~Xiao, M.~Liu, L.~Yuan, H.~Xuan, M.~Valenzuela, X.~S. Chen, X.~Wang, {\em et~al.}, ``Tinyclip: Clip distillation via affinity mimicking and weight inheritance,'' in {\em Proceedings of the IEEE/CVF International Conference on Computer Vision}, pp.~21970--21980, 2023.

\bibitem{li2023inverse}
X.~Li, Z.~Wang, and C.~Xie, ``An inverse scaling law for clip training,'' {\em Advances in Neural Information Processing Systems}, vol.~36, pp.~49068--49087, 2023.

\bibitem{xu2025balancebenchmark}
S.~Xu, M.~Cui, C.~Huang, H.~Wang, and D.~Hu, ``Balancebenchmark: A survey for multimodal imbalance learning,'' {\em arXiv preprint arXiv:2502.10816}, 2025.

\bibitem{yang2024facilitating}
Y.~Yang, F.~Wan, Q.-Y. Jiang, and Y.~Xu, ``Facilitating multimodal classification via dynamically learning modality gap,'' {\em Advances in Neural Information Processing Systems}, vol.~37, pp.~62108--62122, 2024.

\bibitem{liao2025benchmarking}
C.~Liao, K.~Lei, X.~Zheng, J.~Moon, Z.~Wang, Y.~Wang, D.~P. Paudel, L.~Van~Gool, and X.~Hu, ``Benchmarking multi-modal semantic segmentation under sensor failures: Missing and noisy modality robustness,'' {\em arXiv preprint arXiv:2503.18445}, 2025.

\bibitem{leng2024curse}
S.~Leng, Y.~Xing, Z.~Cheng, Y.~Zhou, H.~Zhang, X.~Li, D.~Zhao, S.~Lu, C.~Miao, and L.~Bing, ``The curse of multi-modalities: Evaluating hallucinations of large multimodal models across language, visual, and audio,'' {\em arXiv preprint arXiv:2410.12787}, 2024.

\bibitem{guo2023modality}
Y.~Guo, L.~Nie, H.~Cheng, Z.~Cheng, M.~Kankanhalli, and A.~Del~Bimbo, ``On modality bias recognition and reduction,'' {\em ACM Transactions on Multimedia Computing, Communications and Applications}, vol.~19, no.~3, pp.~1--22, 2023.

\bibitem{vosoughi2024cross}
A.~Vosoughi, S.~Deng, S.~Zhang, Y.~Tian, C.~Xu, and J.~Luo, ``Cross modality bias in visual question answering: A causal view with possible worlds vqa,'' {\em IEEE Transactions on Multimedia}, 2024.

\bibitem{niu2021counterfactual}
Y.~Niu, K.~Tang, H.~Zhang, Z.~Lu, X.-S. Hua, and J.-R. Wen, ``Counterfactual vqa: A cause-effect look at language bias,'' in {\em Proceedings of the IEEE/CVF conference on computer vision and pattern recognition}, pp.~12700--12710, 2021.

\bibitem{gat2020removing}
I.~Gat, I.~Schwartz, A.~Schwing, and T.~Hazan, ``Removing bias in multi-modal classifiers: Regularization by maximizing functional entropies,'' {\em Advances in Neural Information Processing Systems}, vol.~33, pp.~3197--3208, 2020.

\bibitem{ramakrishnan2018overcoming}
S.~Ramakrishnan, A.~Agrawal, and S.~Lee, ``Overcoming language priors in visual question answering with adversarial regularization,'' {\em Advances in neural information processing systems}, vol.~31, 2018.

\bibitem{lee2024vlind}
K.-i. Lee, M.~Kim, S.~Yoon, M.~Kim, D.~Lee, H.~Koh, and K.~Jung, ``Vlind-bench: Measuring language priors in large vision-language models,'' {\em arXiv preprint arXiv:2406.08702}, 2024.

\bibitem{liu2024insight}
X.~Liu, W.~Wang, Y.~Yuan, J.-t. Huang, Q.~Liu, P.~He, and Z.~Tu, ``Insight over sight? exploring the vision-knowledge conflicts in multimodal llms,'' {\em arXiv preprint arXiv:2410.08145}, 2024.

\bibitem{yue2024mmmuv1}
X.~Yue, Y.~Ni, K.~Zhang, T.~Zheng, R.~Liu, G.~Zhang, S.~Stevens, D.~Jiang, W.~Ren, Y.~Sun, {\em et~al.}, ``Mmmu: A massive multi-discipline multimodal understanding and reasoning benchmark for expert agi,'' in {\em Proceedings of the IEEE/CVF Conference on Computer Vision and Pattern Recognition}, pp.~9556--9567, 2024.

\bibitem{zhao2024mmicl}
H.~Zhao, Z.~Cai, S.~Si, X.~Ma, K.~An, L.~Chen, Z.~Liu, S.~Wang, W.~Han, and B.~Chang, ``Mmicl: Empowering vision-language model with multi-modal in-context learning,'' in {\em ICLR}, 2024.

\bibitem{pi2024strengthening}
R.~Pi, T.~Han, W.~Xiong, J.~Zhang, R.~Liu, R.~Pan, and T.~Zhang, ``Strengthening multimodal large language model with bootstrapped preference optimization,'' in {\em European Conference on Computer Vision}, pp.~382--398, Springer, 2024.

\bibitem{liu2024paying}
S.~Liu, K.~Zheng, and W.~Chen, ``Paying more attention to image: A training-free method for alleviating hallucination in lvlms,'' in {\em European Conference on Computer Vision}, pp.~125--140, Springer, 2024.

\bibitem{zhao2024looking}
H.~Zhao, S.~Si, L.~Chen, Y.~Zhang, M.~Sun, M.~Zhang, and B.~Chang, ``Looking beyond text: Reducing language bias in large vision-language models via multimodal dual-attention and soft-image guidance,'' {\em arXiv preprint arXiv:2411.14279}, 2024.

\bibitem{zhang2025debiasing}
Z.~Zhang, H.~Tang, J.~Sheng, Z.~Zhang, Y.~Ren, Z.~Li, D.~Yin, D.~Ma, and T.~Liu, ``Debiasing multimodal large language models via noise-aware preference optimization,'' {\em arXiv preprint arXiv:2503.17928}, 2025.

\bibitem{li2025devil}
Z.~Li, X.~Wen, J.~Lou, Y.~Ji, Y.~Lu, X.~Han, D.~Zhang, and L.~Sun, ``The devil is in the details: Tackling unimodal spurious correlations for generalizable multimodal reward models,'' {\em arXiv preprint arXiv:2503.03122}, 2025.

\bibitem{liu2024deepseek}
A.~Liu, B.~Feng, B.~Xue, B.~Wang, B.~Wu, C.~Lu, C.~Zhao, C.~Deng, C.~Zhang, C.~Ruan, {\em et~al.}, ``Deepseek-v3 technical report,'' {\em arXiv preprint arXiv:2412.19437}, 2024.

\bibitem{bi2024deepseek}
X.~Bi, D.~Chen, G.~Chen, S.~Chen, D.~Dai, C.~Deng, H.~Ding, K.~Dong, Q.~Du, Z.~Fu, {\em et~al.}, ``Deepseek llm: Scaling open-source language models with longtermism,'' {\em arXiv preprint arXiv:2401.02954}, 2024.

\bibitem{naveed2023comprehensive}
H.~Naveed, A.~U. Khan, S.~Qiu, M.~Saqib, S.~Anwar, M.~Usman, N.~Akhtar, N.~Barnes, and A.~Mian, ``A comprehensive overview of large language models,'' {\em arXiv preprint arXiv:2307.06435}, 2023.

\bibitem{ren2023masked}
B.~Ren, Y.~Liu, Y.~Song, W.~Bi, R.~Cucchiara, N.~Sebe, and W.~Wang, ``Masked jigsaw puzzle: A versatile position embedding for vision transformers,'' in {\em Proceedings of the IEEE/CVF Conference on Computer Vision and Pattern Recognition}, pp.~20382--20391, 2023.

\bibitem{han2022survey}
K.~Han, Y.~Wang, H.~Chen, X.~Chen, J.~Guo, Z.~Liu, Y.~Tang, A.~Xiao, C.~Xu, Y.~Xu, {\em et~al.}, ``A survey on vision transformer,'' {\em IEEE transactions on pattern analysis and machine intelligence}, vol.~45, no.~1, pp.~87--110, 2022.

\bibitem{ren2024bringing}
B.~Ren, G.~Mei, D.~P. Paudel, W.~Wang, Y.~Li, M.~Liu, R.~Cucchiara, L.~Van~Gool, and N.~Sebe, ``Bringing masked autoencoders explicit contrastive properties for point cloud self-supervised learning,'' in {\em Proceedings of the Asian Conference on Computer Vision}, pp.~2034--2052, 2024.

\bibitem{liu2021swin}
Z.~Liu, Y.~Lin, Y.~Cao, H.~Hu, Y.~Wei, Z.~Zhang, S.~Lin, and B.~Guo, ``Swin transformer: Hierarchical vision transformer using shifted windows,'' in {\em Proceedings of the IEEE/CVF international conference on computer vision}, pp.~10012--10022, 2021.

\bibitem{ren2024sharing}
B.~Ren, Y.~Li, J.~Liang, R.~Ranjan, M.~Liu, R.~Cucchiara, L.~V. Gool, M.-H. Yang, and N.~Sebe, ``Sharing key semantics in transformer makes efficient image restoration,'' {\em Advances in Neural Information Processing Systems}, vol.~37, pp.~7427--7463, 2024.

\bibitem{huang2024learning}
S.~Huang, B.~Gong, Y.~Feng, X.~Chen, Y.~Fu, Y.~Liu, and D.~Wang, ``Learning disentangled identifiers for action-customized text-to-image generation,'' in {\em Proceedings of the IEEE/CVF Conference on Computer Vision and Pattern Recognition}, pp.~7797--7806, 2024.

\bibitem{languagebind2024}
B.~Zhu, B.~Lin, M.~Ning, Y.~Yan, J.~Cui, H.~Wang, Y.~Pang, W.~Jiang, J.~Zhang, Z.~Li, C.~Zhang, Z.~Li, W.~Liu, and L.~Yuan, ``Languagebind: Extending video-language pretraining to n-modality by language-based semantic alignment,'' in {\em {ICLR}}, OpenReview.net, 2024.

\bibitem{UniBind2024}
Y.~Lyu, X.~Zheng, J.~Zhou, and L.~Wang, ``Unibind: Llm-augmented unified and balanced representation space to bind them all,'' in {\em {CVPR}}, pp.~26742--26752, {IEEE}, 2024.

\bibitem{rahman2016optimizing}
M.~A. Rahman and Y.~Wang, ``Optimizing intersection-over-union in deep neural networks for image segmentation,'' in {\em International symposium on visual computing}, pp.~234--244, Springer, 2016.

\bibitem{rezatofighi2019generalized}
H.~Rezatofighi, N.~Tsoi, J.~Gwak, A.~Sadeghian, I.~Reid, and S.~Savarese, ``Generalized intersection over union: A metric and a loss for bounding box regression,'' in {\em Proceedings of the IEEE/CVF conference on computer vision and pattern recognition}, pp.~658--666, 2019.

\bibitem{hu2020subjective}
B.~Hu, L.~Li, J.~Wu, and J.~Qian, ``Subjective and objective quality assessment for image restoration: A critical survey,'' {\em Signal Processing: Image Communication}, vol.~85, p.~115839, 2020.

\bibitem{xu2024lvlm}
P.~Xu, W.~Shao, K.~Zhang, P.~Gao, S.~Liu, M.~Lei, F.~Meng, S.~Huang, Y.~Qiao, and P.~Luo, ``Lvlm-ehub: A comprehensive evaluation benchmark for large vision-language models,'' {\em IEEE Transactions on Pattern Analysis and Machine Intelligence}, 2024.

\bibitem{liu2024valor}
J.~Liu, S.~Chen, X.~He, L.~Guo, X.~Zhu, W.~Wang, and J.~Tang, ``Valor: Vision-audio-language omni-perception pretraining model and dataset,'' {\em IEEE Transactions on Pattern Analysis and Machine Intelligence}, 2024.

\bibitem{dave2024multimodal}
V.~Dave, F.~Lygerakis, and E.~Rueckert, ``Multimodal visual-tactile representation learning through self-supervised contrastive pre-training,'' in {\em 2024 IEEE International Conference on Robotics and Automation (ICRA)}, pp.~8013--8020, IEEE, 2024.

\bibitem{lyu2024omnibind}
Y.~Lyu, X.~Zheng, D.~Kim, and L.~Wang, ``Omnibind: Teach to build unequal-scale modality interaction for omni-bind of all,'' {\em arXiv preprint arXiv:2405.16108}, 2024.

\bibitem{zhang2025soft}
N.~Zhang, J.~Ren, Y.~Dong, X.~Yang, R.~Bian, J.~Li, G.~Gu, and X.~Zhu, ``Soft robotic hand with tactile palm-finger coordination,'' {\em Nature Communications}, vol.~16, no.~1, p.~2395, 2025.

\bibitem{kirschner2025categorizing}
R.~J. Kirschner, K.~Karacan, A.~Melone, and S.~Haddadin, ``Categorizing robots by performance fitness into the tree of robots,'' {\em Nature Machine Intelligence}, pp.~1--12, 2025.

\bibitem{agarwal2025vision}
A.~Agarwal, A.~Wilson, T.~Man, E.~Adelson, I.~Gkioulekas, and W.~Yuan, ``Vision-based tactile sensor design using physically based rendering,'' {\em Communications Engineering}, vol.~4, no.~1, p.~21, 2025.

\bibitem{gbagbe2024bi}
K.~F. Gbagbe, M.~A. Cabrera, A.~Alabbas, O.~Alyunes, A.~Lykov, and D.~Tsetserukou, ``Bi-vla: Vision-language-action model-based system for bimanual robotic dexterous manipulations,'' in {\em 2024 IEEE International Conference on Systems, Man, and Cybernetics (SMC)}, pp.~2864--2869, IEEE, 2024.

\bibitem{bennetot2024practical}
A.~Bennetot, I.~Donadello, A.~El~Qadi El~Haouari, M.~Dragoni, T.~Frossard, B.~Wagner, A.~Sarranti, S.~Tulli, M.~Trocan, R.~Chatila, {\em et~al.}, ``A practical tutorial on explainable ai techniques,'' {\em ACM Computing Surveys}, vol.~57, no.~2, pp.~1--44, 2024.

\bibitem{dwivedi2023explainable}
R.~Dwivedi, D.~Dave, H.~Naik, S.~Singhal, R.~Omer, P.~Patel, B.~Qian, Z.~Wen, T.~Shah, G.~Morgan, {\em et~al.}, ``Explainable ai (xai): Core ideas, techniques, and solutions,'' {\em ACM Computing Surveys}, vol.~55, no.~9, pp.~1--33, 2023.

\end{thebibliography}

\end{document}